\documentclass[sigconf]{acmart}
\usepackage{multirow}
\usepackage{graphicx}
\usepackage{makecell}
\AtBeginDocument{%
  }

\begin{document}

\title{Peek into the `White-Box': A Field Study on Bystander Engagement with Urban Robot Uncertainty}

\author{Xinyan Yu}
\email{xinyan.yu@sydney.edu.au}
\orcid{0000-0001-8299-3381}
\affiliation{Design Lab, Sydney School of Architecture, Design and Planning
  \institution{The University of Sydney}
  \city{Sydney}
  \state{NSW}
  \country{Australia}
}
\author{Marius Hoggenmueller}
\email{marius.hoggenmueller@sydney.edu.au}
\orcid{0000-0002-8893-5729}
\affiliation{Design Lab, Sydney School of Architecture, Design and Planning
  \institution{The University of Sydney} 
  \city{Sydney}
  \state{NSW}
  \country{Australia}
}

\author{Tram Thi Minh Tran}
\email{tram.tran@sydney.edu.au}
\orcid{0000-0002-4958-2465}
\affiliation{Design Lab, Sydney School of Architecture, Design and Planning
  \institution{The University of Sydney}
  \city{Sydney}
  \state{NSW}
  \country{Australia}
}

\author{Yiyuan Wang}
\email{yiyuan.wang@sydney.edu.au}
\orcid{0000-0003-2610-1283}
\affiliation{Design Lab, Sydney School of Architecture, Design and Planning
  \institution{The University of Sydney}
  \city{Sydney}
  \state{NSW}
  \country{Australia}
}

\author{Qiuming Zhang}
\email{qzha4601@uni.sydney.edu.au}
\orcid{1234-5678-9012}
\affiliation{Design Lab, Sydney School of Architecture, Design and Planning
  \institution{The University of Sydney}
  \city{Sydney}
  \state{NSW}
  \country{Australia}
}

\author{Martin Tomitsch}
\email{Martin.Tomitsch@uts.edu.au}
\orcid{0000-0003-1998-2975}
\affiliation{Transdisciplinary School,
  \institution{University of Technology Sydney}
  \city{Sydney}
  \state{NSW}
  \country{Australia}
}

\renewcommand{\shortauthors}{Yu et al.}

\begin{abstract}
Uncertainty inherently exists in the autonomous decision-making process of robots. Involving humans in resolving this uncertainty not only helps robots mitigate it but is also crucial for improving human-robot interactions. However, in public urban spaces filled with unpredictability, robots often face heightened uncertainty without direct human collaborators. 
This study investigates how robots can engage bystanders for assistance in public spaces when encountering uncertainty and examines how these interactions impact bystanders' perceptions and attitudes towards robots. We designed and tested a speculative `peephole' concept that engages bystanders in resolving urban robot uncertainty. Our design is guided by considerations of non-intrusiveness and eliciting initiative in an implicit manner, considering bystanders' unique role as non-obligated participants in relation to urban robots. Drawing from field study findings, we highlight the potential of involving bystanders to mitigate urban robots' technological imperfections to both address operational challenges and foster public acceptance of urban robots. Furthermore, we offer design implications to encourage bystanders' involvement in mitigating the imperfections.
\end{abstract}




\received{20 February 2007}
\received[revised]{12 March 2009}
\received[accepted]{5 June 2009}

\maketitle

\section{Introduction}
Robot's autonomous decision-making relies on the acquisition and analysis of environmental data, a process inherently subject to uncertainty from various sources, such as sensor noise and inaccuracies in machine learning models~\cite{Loquerci2020UncertaintyFramework}. Integrating human perception and cognition into robot decision-making can effectively address such uncertainty, and is also essential for facilitating smoother human-robot interactions and fostering trust~\cite{leusmann2023UncertaintyLoop, SarahCHI2024Uncertainty}.

The uncertainty challenges become more pronounced when robots transition from relatively static and controlled environments to public urban spaces. These contexts are characterised by increased levels of dynamics and unpredictability~\cite{Putten2020Forgotten,Xinyan2024OutofPlace,Weinberg2023SharingSidewalk}, and urban robots often operate unsupervised, lacking human collaborators to turn to when uncertainty arises. This absence of support can lead to undesired decision-making, resulting in negative outcomes that not only hinder operational effectiveness but also undermine public perception of these technologies. 
A notable example occurred in September 2022, when a delivery robot, after a moment of hesitation, decided to cross into a crime scene, blatantly ignoring police tape and leaving onlookers confused and surprised. The incident was captured on video\footnote{Available at \url{https://www.vice.com/en/article/93adae/food-delivery-robot-casually-drives-under-police-tape-through-active-crime-scene}.} and quickly went viral on social media, sparking public concerns about the reliability of the technology. 

As robots become increasingly prevalent in public urban spaces, bystanders---defined as \emph{`incidentally copresent persons'} (InCoPs)~\cite{Putten2020Forgotten}, individuals who do not have any prior intentions to engage with the robot---are gaining attention in human-robot interactions. Casual collaboration between bystanders and robots can emerge, particularly in situations where the robot requires assistance~\cite{Xinyan2024CasualCollaboration,Chi2024ShouldIHelp,Holm2022Signaling, Pelikan2024Encounter,Weinberg2023SharingSidewalk}. These emerging participants in HRI have the potential to assist urban robots in resolving uncertainty. Therefore, it is crucial to investigate how to engage bystanders in addressing such robot uncertainty, as well as how these interactions shape bystanders' perceptions and attitudes towards robots.

Distinct from traditional human-robot collaboration settings, the engagement of bystanders in casual collaborations requires tailored interaction strategies. Unlike traditional collaborators, who have a pre-established collaborative relationship and shared task goals, bystanders have no obligation to engage with the robot. Thus, engaging bystanders should consider minimising intrusiveness while fostering their self-initiative, rather than creating a sense of obligation, which can often result from explicit, direct verbal prompts~\cite{schneider2021step,Yu2024Playful}. To address this, we developed a speculative \emph{peephole} design concept that draws on the considerations of non-intrusiveness and implicit initiative-eliciting. Inspired by the principles of ludic engagement---\emph{`activities motivated by curiosity, exploration, and reflection rather than externally defined tasks'}~\cite{Gaver2004DriftTable}---this concept conceals the robot's uncertainty information behind a pair of binocular scopes, which open only when uncertainty arises. Rather than exposing and broadcasting robot uncertainty information through explicit modes of communication (e.g.,~visual displays or spoken language), the intent of the concept is to intuitively elicit bystanders' curiosity and encourage them to peek at hidden information. The peephole transforms direct help requests into a self-initiated, playful discovery process while reducing the persuasiveness and minimising the disruption that direct help requests could cause.

We built a mobile robot resembling a typical sidewalk service robot (e.g., delivery robot) to implement this concept and probed it in a Wizard of Oz field study~\cite{dahlback1993wizard}, in order to investigate bystanders' spontaneous reactions to the non-intrusive and initiative-eliciting engagement strategy. The robot was deliberately staged to get stuck in scenarios where it faced uncertainty, i.e., encountering ambiguous obstacles in its path. The field study was conducted across three different locations over a period of nine days in total, where we observed bystander reactions and conducted interviews. The design artefact evolved throughout the study, informed by on-site observations and reflections. Our work aims to investigate how bystanders respond when robots engage them in situations of uncertainty and examine the impact of such engagement on people's perceptions and attitudes towards robots. Additionally, we seek to uncover the multifaceted nature of employing implicit and non-intrusive strategies to engage bystanders in public spaces.

%[Contribution Statement]
Our work contributes to the fields of HCI and interaction design by: (1) developing a speculative design concept that engages bystanders in resolving urban robots’ uncertainties and documenting the artefact’s evolution through a field study; (2) providing empirical insights into the potential of involving bystanders to mitigate urban robots' imperfections, not only addressing urban robots' operational challenges but also fostering their public acceptance; (3) offering design implications for implicit and non-intrusive engagement strategies that encourage casual human-robot collaboration. Our investigation aligns with emerging design research perspectives~\cite{lupetti2019citizenship, Kuijer2018Coperformance, marenko2016animistic}, which increasingly view human-robot interaction as a symbiotic relationship rather than focusing solely on robots' independent capabilities.

\section{Related Work}
\subsection{Integrating human autonomy in AI}
As AI systems become increasingly sophisticated and prevalent, concerns about their accuracy, fairness, and ethical implications have grown. Maintaining human autonomy in the loop of AI decision-making has emerged as a crucial strategy to address these issues~\cite{zanzotto2019human,WU2022SurveyHITL,shneiderman2022human}. In this paradigm, humans have the opportunity to evaluate (e.g., loan decisions or crime judgements made by AI~\cite{nakao2022Fairness, agudo2024impact}), modify (e.g., robot vision recognition results ~\cite{Cai2021CV,Abraham2021AdaptiveAutonomy}), and contest (e.g., decisions made by public AI systems~\cite{Alfrink2023ContestableAI}) the results produced by AI systems, particularly when the system has uncertainty in its decisions~\cite{Abraham2021AdaptiveAutonomy,Malinin2018PredictiveUncertainty,Charles2019Confidence}. For example, ~\cite{Abraham2021AdaptiveAutonomy} proposed a vision-based robotic system with adaptive autonomy, which temporarily lowers its autonomy level to involve human operators in decision-making when the reliability of its computer vision model is compromised. In addition to enhancing AI decision-making results, a growing body of research is advocating for interactive machine learning, which actively engages humans in the learning processes of machine learning models~\cite{fails2003interactiveLearning}, further centering human autonomy in the development of AI systems.

While the involvement of humans in AI has traditionally focused on experts with professional knowledge, researchers have increasingly emphasised the inclusion of non-experts~\cite{Ramos2020InteractiveMachineTeaching}. The HCI community has also begun to respond to this emerging paradigm by exploring the development of interaction strategies that enable non-experts to actively participate in the AI decision-making and development process~\cite{Yang2018GIML,Feng2023UXIML,nakao2022Fairness}.

\subsection{Robot uncertainty and human intervention}
Robot's autonomous decision-making relies on the acquisition and analysis of environmental data, during which inherent uncertainties can arise from various sources, such as sensor noise and inaccuracies in machine learning models~\cite{Loquerci2020UncertaintyFramework}. In human-robot collaborative tasks, it is essential for robots to communicate uncertainties to their human collaborators, enabling humans to better understand the robots' operations and address these uncertainties in their responses, thereby fostering optimised collaboration and enhancing trust~\cite{SarahCHI2024Uncertainty,leusmann2023UncertaintyLoop,Hough2017RobotUncertainy}. Additionally, informing humans of uncertainties allows robots to benefit from human-provided information that helps to reduce these uncertainties~\cite{Shin2023Uncertainty-Resolving,Abraham2021AdaptiveAutonomy,SarahCHI2024Uncertainty}. To communicate uncertainty to human collaborators, previous research has explored the use of verbal~\cite{Shin2023Uncertainty-Resolving} and non-verbal cues, such as motion speed~\cite{Hough2017RobotUncertainy}, hesitation gestures~\cite{Moon2021Hesitation}, and graphic visualisation~\cite{SarahCHI2024Uncertainty}.  

Shifting from relatively static and controlled human-robot collaborative settings~(e.g., labs, factories), robots are now operating in public urban spaces characterised by increasing uncertainty due to their unpredictable and ever-changing nature~\cite{Alatise2020ReviewChallenge,Ferrer2013Social-awareRobotNavigation}. However, urban robots often operate unsupervised, without human collaborators to rely on when uncertainty arises. In such instances, bystanders could serve as a source of assistance, yet urban robots' uncertainty always remains hidden, preventing bystanders from engaging in addressing these challenges. While there is no research directly investigating the engagement of bystanders in addressing robot uncertainty in public spaces, several notable precedents exploring bystander input to supplement robotic autonomy highlight its potential. For instance, a speculative project involving a wandering robot, Tweenbot, devoid of sensors and autonomous navigation, successfully relied on the assistance of passersby to reach its destination~\cite{tweenbots}. Similarly, the Autonomous City Explorer (ACE) navigated urban spaces by soliciting directions from pedestrians, foregoing the use of maps or GPS systems~\cite{Weiss2010ACE}. 

\subsection{Casual human-robot collaboration}
With robots transitioning from controlled to uncontrolled environments, casual collaborations---distinguished from the planned and anticipated human-robot collaborations between robots and human teammates~\cite{Cila2022HumanAgentCollaboration}---are increasingly happening spontaneously between robots and bystanders in public spaces. Recent field observation studies have witnessed such casual collaboration in various situations where urban robots are in need of assistance. For example, pedestrians have voluntarily assisted immobilised robots by removing obstacles~\cite{Weinberg2023SharingSidewalk}, giving delivery robots struggling in heavy snowfall a gentle push~\cite{Dobrosovestnova2022LittleHelp}, or even pausing their ongoing work to allow delivery robots to pass through~\cite{Pelikan2024Encounter}. 

In contrast to traditional human-robot collaboration, the dynamics of casual collaboration shift significantly due to misaligned task objectives and their role as non-obligated participants in relation to urban robots. Additionally, contextual factors such as task scenarios and the bystander's current activity~\cite{Fischer2014SocialFraming, Huttenrauch2003ToHelporNot}, the robot's perceived legitimacy and risk~\cite{Booth2017Piggy}, along with subjective factors like trust and perceived robot competence~\cite{Cameron2015Buttons}, collectively influence the formation of such casual collaboration. Furthermore, \citet{hakli2023helpingAsWorkCare} brought forth a nuanced discussion on the complexities of bystanders assisting commercially deployed robots in public spaces. They use the notion of ambiguity as a productive lens to shed light on different perspectives in robot-helping situations, highlighting concerns around invisible labour but also acknowledging the relational and affective dimensions of these interactions.

Thus, there is a need for tailored design strategies to effectively engage bystanders in casual collaboration with robots. However, both academic research~\cite{Fischer2014SocialFraming,Liang2023AskForDirections,Rosenthal2012OfficeHelp, Wullschleger2002paradox} and commercially deployed robots~\cite{Boos2022Polite} predominantly rely on verbal communication to facilitate these interactions, with limited exploration of using other less intrusive approaches that suited to the bystander’s non-obligated nature~\cite{Holm2022Signaling}. 
Recognising this gap, \citet{Xinyan2024CasualCollaboration} explored the use of non-verbal strategies to foster casual human-robot collaboration and highlighted the potential of leveraging the inherent expressiveness in the functional aspects or form of robots. They emphasise the need for implicit, non-persuasive approaches that frame bystander assistance as a voluntary, spontaneous act rather than an obligation. Their subsequent study explored game-inspired concepts as robot help-seeking strategies~\cite{Yu2024Playful}. They found that robot help-seeking through verbal speech was perceived as impolite by bystanders, underscoring the potential of curating humans' innate playfulness as an intrinsic motivation to foster casual collaboration, rather than relying on explicit help requests. While providing insights into engaging bystanders in an implicit manner, these findings were derived from controlled lab settings, where spontaneous reactions from bystanders could not be observed in a fully natural context.

\section{Field Study}
Involving human autonomy can enhance AI decision-making, particularly in scenarios where system confidence is low, as exemplified by robots facing uncertainty. While prior research has explored engaging human collaborators in mitigating robot uncertainty---demonstrating benefits such as optimised operation and enhanced trust---the dynamics of engagement between bystanders and urban robots remain unexplored. Drawing on insights from previous work on strategies for casual human-robot collaboration, we
developed a speculative design concept and probed it in a field study to investigate bystanders’ reactions in real-world settings and assess how such engagement shapes public perceptions and attitudes
towards robots.
\begin{figure*}[h]
\begin{center}
\includegraphics[width=\textwidth]{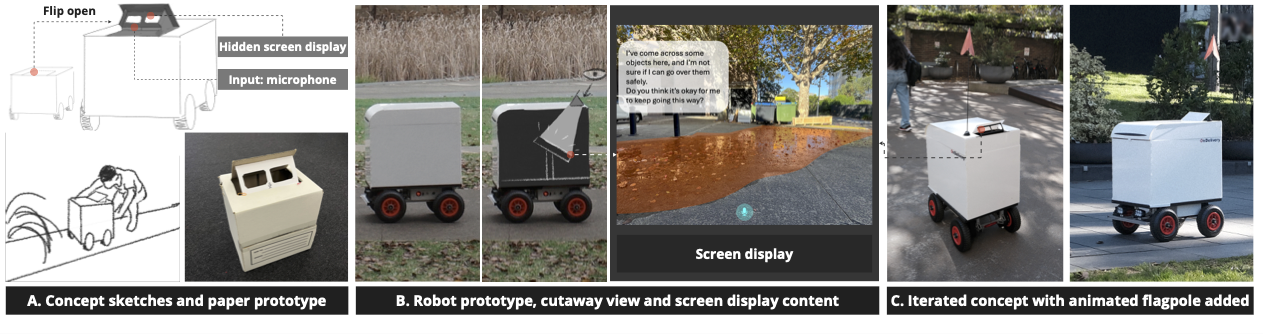}
\end{center}
\vspace{-8pt} %
\caption{Concept development from initial sketches to final implementation: (A) Concept sketches, interaction illustrations, and a paper prototype in the early stage; (B) Robot prototype during the initial deployment, with a cutaway view and screen display content; (C) The iterated design concept featuring an added animated flagpole.}
\label{Concept}
\Description{Figure 1: This figure illustrates the concept development process for our prototype, from initial sketches to the final implementation across three stages, displayed as three subfigures from left (A) to right (B): (A) Concept sketches and paper prototype: This stage shows the initial design sketches of the robot. The sketches depict a box-shaped robot featuring a pair of flip-open binocular-like lenses, a hidden screen display, and a microphone input. Accompanying the sketches is a physical paper prototype made from a white cardboard box. The box includes a pair of binocular-like lenses cut out from cardboard, located at the rear top of the box, serving as the peephole; (B) Robot prototype, cutaway view, and screen display content: This subfigure presents the robot prototype during its initial deployment. The robot has a rectangular, box-like shape with rounded front edges and a primarily white exterior. The image includes a cutaway view, revealing internal components such as the screen. The screen displays a message asking for assistance, illustrating the robot's uncertainty. The displayed message reads: "I've come across some debris here, and I'm not sure if I can go over them safely. Do you think it's okay for me to keep going this way?"; (C) Final robot prototype with animated flagpole added: In the final stage, the robot is presented with the addition of an animated flagpole designed. The flagpole consists of a black plastic stick with a red flag at its top, and it is mounted on the top surface of the robot at the rear side. This subfigure includes photos of the robot captured from two angles, a front-side view and a rear-side view.}
\end{figure*}
\subsection{Design concept}

\subsubsection{Design rationales}
Informed by both theoretical underpinnings and practical design research investigations on robot help-seeking from bystanders (detailed in Section 2), our design was initially guided by two key considerations: (1) the initiative should come from the bystanders themselves, without being prompted by explicit verbal requests that could create a sense of obligation, and (2) the design should minimise disruption to the surrounding environment and the activities bystanders are currently engaged in.

To this end, our design draws inspiration from ludic engagement~\cite{Gaver2004DriftTable}, which emphasises activities motivated by curiosity, exploration, and reflection rather than externally-defined tasks. This approach-initially introduced in HCI and interaction design research-has since been extended to the domain of HRI~\cite{Lee2020LudicHRI,Lee2019Bubble,Marius2020Woody}. For example, ~\citet{Lee2019Bubble} presented a case study where humans' inherent playfulness was leveraged by a bubble-bursting robot to captivate people in public space and create enjoyable experiences. The ability of ludic engagement to spark curiosity and promote interactions among passersby aligns with our goal of encouraging bystander-initiated involvement and has the potential to foster positive and enjoyable experiences. Additionally, the exploratory nature of ludic engagement and its openness addresses the ambiguities of robot help-seeking from bystanders~\cite{hakli2023helpingAsWorkCare}, allowing bystander reactions to be curated in an unstructured and spontaneous manner.

\subsubsection{The peephole concept}

The universal metaphor of an AI system as a `black-box,' indicating its opaque internal processes, serves as the starting point for our design. Our goal is to allow bystanders to access the information concealed within this `black-box' when robot uncertainty occurs. This concept aligns well with the interactive strategy of \emph{peephole} to create engaging interactions, as proposed in~\cite{dalsgaard2009peepholes}. This strategy leverages the tension between hidden and revealed information to foster engagement, a technique successfully used in museums to encourage visitors to explore cultural and natural history exhibits~\cite{Cassinelli2005Projector, Edmonds2006CreativeEngagement}. The curiosity-driven nature of \emph{peephole} interactions embodies the exploratory and unstructured essence of ludic design, aligning well with our aim to create engagement that elicits initiative from bystanders while minimising intrusiveness.

Our concept involves a pair of binocular-like lenses that flip open when the robot encounters uncertainties~(see Fig.\ref{Concept}, A). Information about the uncertainty and a request for assistance are displayed on screens hidden behind these lenses. This setup leverages the human instinct of curiosity---`peeking'---to motivate bystander engagement. Bystanders need to bend over to view the information, which is designed to bring them to the same level and perspective as the robot. This approach fluently allows bystanders to assess the environment from the robot's point of view, making it easier for them to understand the situation and offer help.

The displayed content includes a colour overlay on the objects that the robot is uncertain about recognising and making decisions on (e.g., a puddle of water with leaves that cast reflective light, which can be challenging for computer vision to interpret). The image with the colour overlay is accompanied by a textual description of the situation and a request for help~(see Fig.\ref{Concept}, B). Our uncertainty visualisation is similar to the visualisation approach in~\cite{Colley2021SemanticSegmentation}, where the internal processes of autonomous vehicles were communicated to passengers through the semantic segmentation visualisation by highlighting recognised objects.

In terms of input, we opted for an unstructured approach by allowing bystanders to respond freely via speech, using a microphone to capture their input. The decision to use speech as the input rather than selecting different options (e.g., via buttons) was made to encourage more open-ended responses. 

\subsubsection{Concept evolving during deployment}
As the field study progressed, we noticed a low percentage of involvement during the initial days of deployment, with the majority of passersby walking past the robot without paying much attention to it. Thus, as part of our iterative design process, we decided to add additional elements that can attract bystanders' attention to increase engagement. This ongoing development of the design concept aligns with the Research through Design (RtD) approach ~\cite{Zimmerman2007RtD} that we followed, which emphasises the iterative, exploratory process where insights continuously emerge and shape the research.

Prior research has highlighted the human tendency to perceive non-verbal gestures from non-humanoid robotic objects as social signals~\cite{Novikova2014DesignModel,Erel2022emotionSupport, Erel2024OpeningEncounter,Press2022HumorousGestures}, sometimes inviting further interactions~\cite{ju2009approachability,sirkin2015Ottoman}. Social meanings can be attributed to a robot's gestures, even when they are minimal and abstract. For instance, the \emph{Greeting Machine}, an abstract non-humanoid robot, was designed to signal positive and negative social cues during open encounters. In this concept, a small ball moving forward on a dome was perceived as a willingness to interact, while backward movement signalled reluctance~\cite{Anderson-Bashan2018GreetingMachine}.
Drawing inspiration from such effective use of gestural cues to engage interaction in an abstract and minimal manner, and to maintain consistency with the typical design of commercial delivery robots, we added an animated flagpole, a component commonly seen on such robots. The flagpole's animation was designed to wave to attract attention, rotate to indicate a search for help, and repeatedly point in a specific direction, mimicking nodding to directly address an individual.

After another period of deployment without achieving a significant increase in engagement, we further tested whether adding auditory cues could more effectively capture attention from passersby. We introduced a two-note beep, beginning with a higher pitch and followed by a lower one. This sound was chosen because it is interpreted as `negative' in musical terms, potentially signalling something that requires attention, a strategy often used to raise alerts among people~\cite{Liam2020SoundAlarm} and has been used in robot help-seeking~\cite{Holm2022Signaling}.

\subsubsection{Implementation}
We build a mobile robot to implement the concept. The robot's shell is constructed from white acrylic with a curved 3D-printed front, resembling a typical delivery robot (see Fig. \ref{Concept}, C). Its dimensions, approximately 40 cm wide, 60 cm long, and 60 cm high, are comparable to those of a standard sidewalk delivery robot. To ensure passersby identified it as a delivery robot rather than a research prototype, we added an `on delivery' sign to the robot. The robot is powered by a four-wheeled robotic base that can be remotely controlled via a gamepad. The flip-open lenses are operated by a stepper motor, which is controlled by an Arduino board equipped with a Wi-Fi module for remote control. The flagpole is animated by a set of two-dimensional servos, allowing 2-degree of freedom rotation across two different axes, which is also remotely controlled via a gamepad. Finally, the auditory cue is played through a Bluetooth speaker placed inside the robot. 

\subsection{Wizard of Oz set-up}
%wizard 
\begin{figure}[h]
\begin{center}
\includegraphics[width=0.5\textwidth]{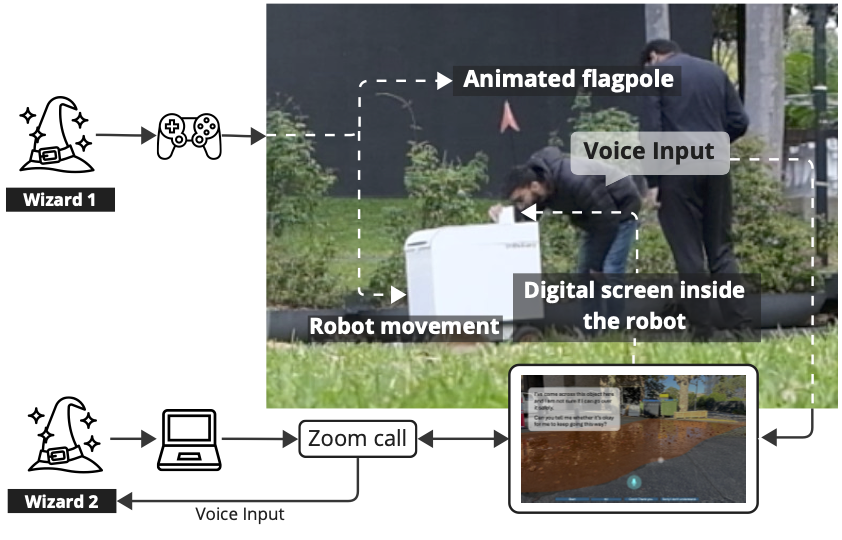}
\end{center}
\vspace{-8pt} %
\caption{Wizard of Oz set-up overview}\label{Setup}
\Description{This figure illustrates the Wizard of Oz setup, where two human operators (wizards) remotely control the robot’s behaviour. The figure overlays a real-world photograph of the robot in action with a diagram that explains the setup. The background shows the robot deployed in an outdoor environment, interacting with bystanders. Key features are labeled directly on the photograph: (1) Animated flagpole: A black stick with a red flag at the rear top of the robot; (2）Voice input: Demonstrates verbal communication between the robot (via Wizard 2) and bystanders;（3）Digital screen: Positioned inside the robot, displaying a message asking for assistance. The displayed text reads: "I've come across some debris here, and I'm not sure if I can go over them safely. Do you think it's okay for me to keep going this way?";（4）Robot movement: Indicates that the robot's movement is controlled remotely by Wizard 1. Overlaying the photograph, the diagram visually represents the Wizard of Oz setup. Wizard 1: Shown with a wizard hat icon, controlling the robot's movement and animated flagpole using a game controller. Wizard 2: Represented by another wizard hat icon, connected to the robot via a Zoom call on a laptop. The arrows in the diagram illustrate the flow of control: Wizard 1 manages physical robot movement and flagpole animation, while Wizard 2 communicates with bystanders' voice input by displaying responses on screen.}
\end{figure}

The field study followed the Wizard of Oz method~\cite{dahlback1993wizard}, wherein two researchers acted as wizards to control the robot's movement, open the lenses, and animate the flag (an overview of the Wizard of Oz setup is shown in Fig.~\ref{Setup}). Wizard 1 remotely drove the robot from the starting point to the location where it encountered uncertainty. Once bystanders intervened and helped resolve the issue, Wizard 1 restored the robot's movement. After each instance of successful engagement and the subsequent interview, Wizard 1 drove the robot back to the starting point and paused for a while before returning to the interaction location where uncertainty was staged. This was intended to ensure that the following passersby had not witnessed the previous interaction and interview. Wizard 1 also controlled the animated flagpole and peephole using a gamepad. Wizard 1 was given instructions on pre-defined motions, including rotating to search for help, waving to attract attention, and nodding to address passersby. In addition, Wizard 1 had the flexibility to adjust or expand these motions in response to bystander interactions.

Wizard 2 monitored what bystanders said in response to the robot via a Zoom call between a laptop and the iPad inside the robot. In response to what passersby said to the robot, Wizard 2 remote-controlled the screen inside by manipulating the displayed text with several pre-determined phrases. For example, phrases like \emph{`Sorry, I don’t understand. Can you repeat?'} were used to elicit clearer instructions when bystanders' utterances were unclear, while \emph{`Thank you'} was displayed in response to bystanders' assistance. This was achieved by manipulating slides over the shared Zoom screen, simulating a conversational interface. 

\subsection{Location and deployment duration}
The field study was conducted in three locations across our university campus, including (1) a sidewalk in front of a faculty building situated at the corner of the university campus, (2) a main pedestrian path that connects public transportation to several major university facilities, and (3) a courtyard inside a faculty building that leads to the main street outside of campus. The first two locations shared similar characteristics, with passersby primarily in commute mode, while in the third location, people were more likely to dwell and engage in social activities. We staged uncertainty by placing a puddle of water with leaves in the robot’s path, where the reflective light could confuse the computer vision system and complicate decision-making, even though it was safe to proceed.

The robot was trialled for approximately 6.5 hours over five weekdays at the first location, 5.5 hours over three weekdays at the second location, and 3.5 hours over three weekdays at the third location, totalling 15.5 hours of deployment (see Table \ref{Overview}). All sessions took place during the semester, ensuring that the flow of people at each location remained roughly consistent across the deployment days.

\begin{table*}[h!]
\centering
\renewcommand{\arraystretch}{0.3}
\small
\caption{Overview of Study Locations and Number of Engagements}
\Description{This table provides an overview of the study locations, durations, and engagement counts across different iterations of the study. The first column describes the three study locations. Location 1 is a sidewalk in front of a faculty building situated at the corner of the university campus, offering moderate foot traffic averaging 7.3 people per minute. Location 2 is a main pedestrian path that connects public transportation to several major university facilities and has the highest average foot traffic of 28.2 people per minute. Location 3 is a courtyard inside a faculty building leading to the main street outside the campus, with the lowest average foot traffic of 5 people per minute.

The second column presents the duration of each study iteration. At Location 1, the study lasted 3.5 hours for Iteration 1 and 3 hours for Iteration 2. At Location 2, the study lasted 2.5 hours for Iteration 2 and 3 hours for Iteration 3. At Location 3, only Iteration 2 was conducted, lasting 3.5 hours.

The third column shows the engagement count for each iteration at each location. Location 1 had the highest engagement count in Iteration 1 with 8 engagements and 6 engagements in Iteration 2. Location 2 recorded 4 engagements during Iteration 2 and 6 engagements during Iteration 3. Location 3, where only Iteration 2 was conducted, recorded 5 engagements.

Footnotes explain that the average foot traffic was manually counted from video footage for 15-minute periods. Additionally, the study's three iterations varied in design: Iteration 1 featured the initial peephole concept, Iteration 2 added an animated flagpole, and Iteration 3 included both the animated flagpole and an auditory cue. This table highlights how location and study iterations influenced engagement levels.}
\label{Overview}
\begin{tabular}{m{2.5cm} m{6cm} m{2.5cm} m{2.5cm}}
\toprule
\textbf{Study Location} & \textbf{Description} & \textbf{Duration} & \textbf{Engagement Count} \\
\midrule
\makecell{\includegraphics[width=2.5cm]{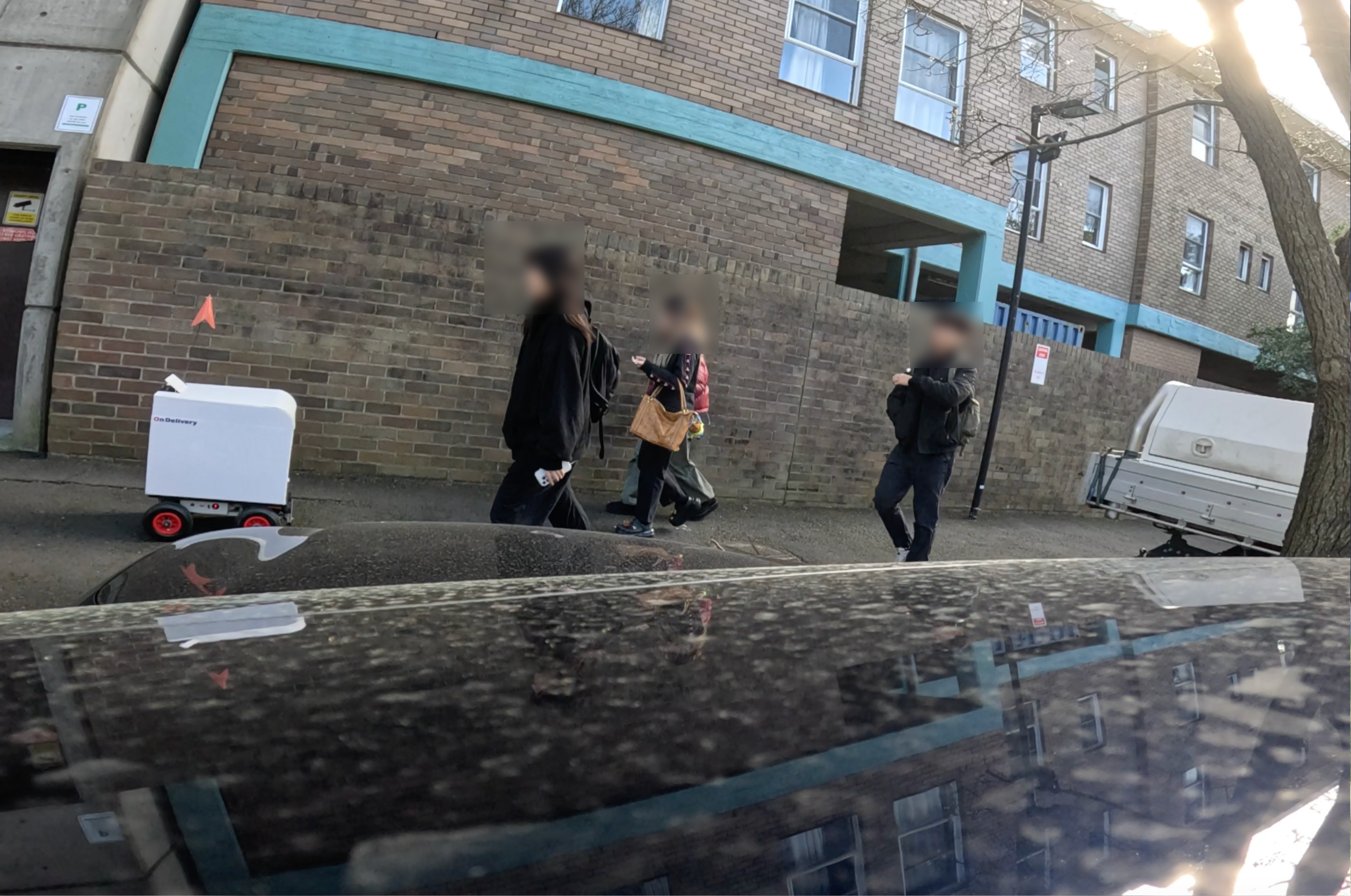}} & 
\makecell{\parbox{6cm}{\centering 
\textbf{Location 1}: A sidewalk in front of a faculty building situated at the corner of the university campus \\ 
\textbf{Average Foot Traffic*}: 7.3/min}} & 
\makecell{\textbf{3.5hrs} (Iteration 1) \\ \textbf{3hrs} (Iteration 2)} & 
\makecell{\textbf{8} (Iteration 1) \\ \textbf{6} (Iteration 2)} \\
\hline
\makecell{\includegraphics[width=2.5cm]{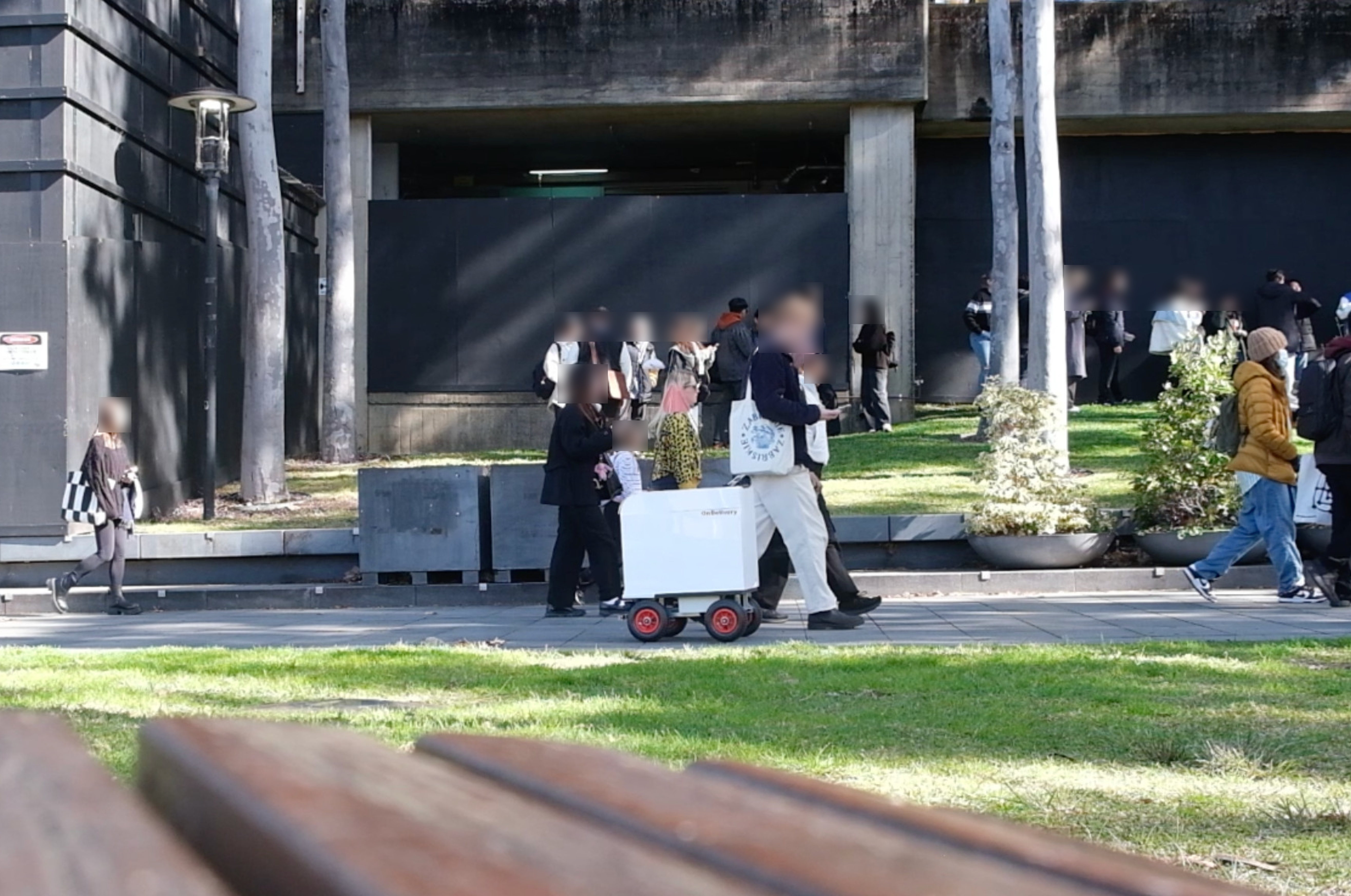}} & 
\makecell{\parbox{6cm}{\centering 
\textbf{Location 2}: A main pedestrian path that connects public transportation to several major university facilities \\ 
\textbf{Average Foot Traffic*}: 28.2/min}} & 
\makecell{\textbf{2.5hrs} (Iteration 2) \\ \textbf{3hrs} (Iteration 3)} & 
\makecell{\textbf{4} (Iteration 2) \\ \textbf{6} (Iteration 3)} \\
\hline
\makecell{\includegraphics[width=2.5cm]{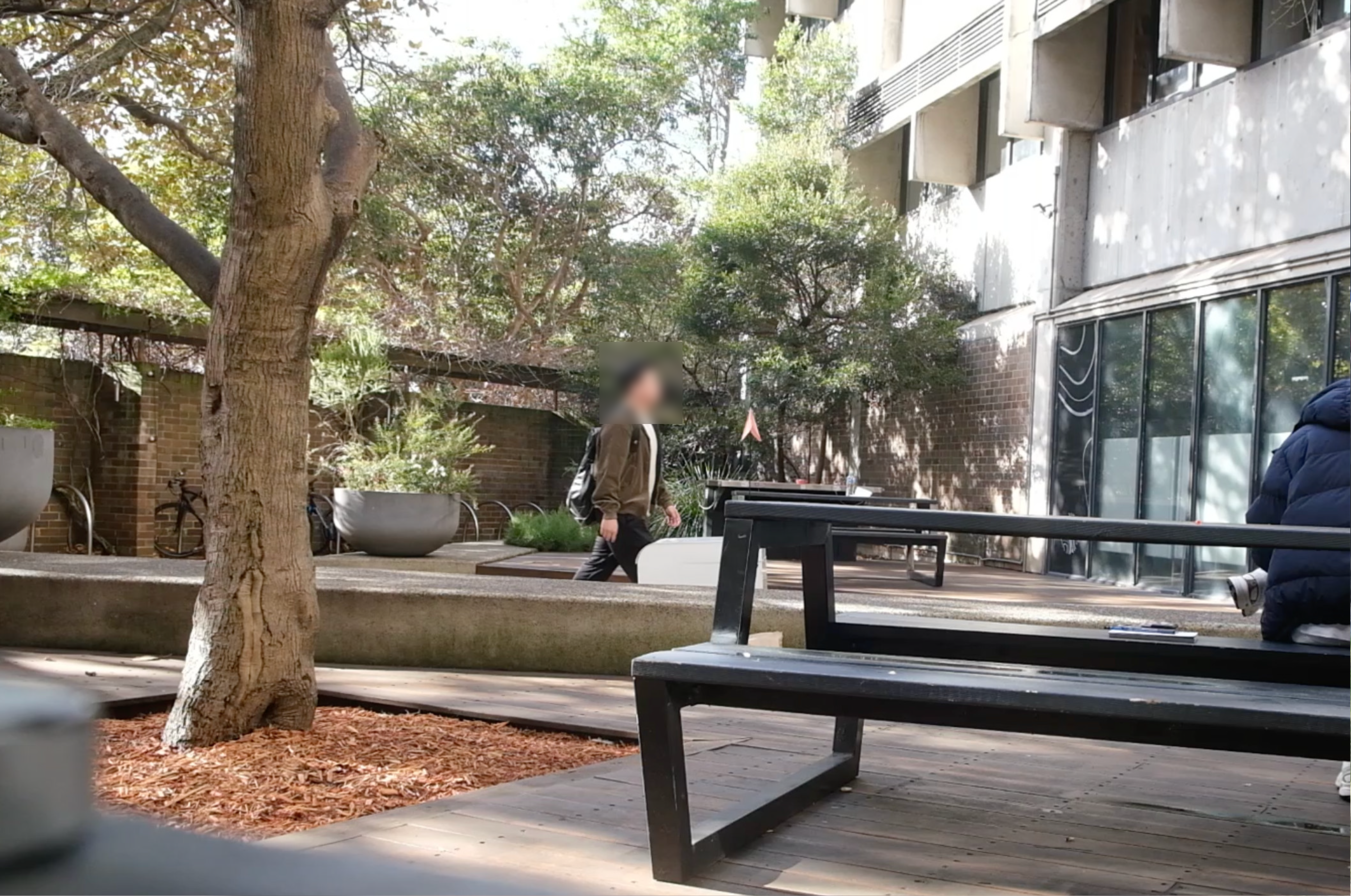}} & 
\makecell{\parbox{6cm}{\centering 
\textbf{Location 3}: A courtyard inside a faculty building that leads to the main street outside of campus \\ 
\textbf{Average Foot Traffic*}: 5/min}} & 
\makecell{\textbf{3.5hrs} (Iteration 2)} & 
\makecell{\textbf{5} (Iteration 2)} \\
\bottomrule
\end{tabular}

\vspace{5pt} % Adds a little space before the footnotes

\parbox{0.8\linewidth}{  % Make footnotes the same width as the table
    \footnotesize
    \noindent * Average foot traffic is estimated by manually counting people passing by in the video over several periods, totaling 15 minutes for each location.\\
    ** Iteration 1: the initial peephole concept; Iteration 2: peephole concept with animated flagpole added; Iteration 3: peephole concept with both animated flagpole and auditory cue added.
}

\end{table*}

\subsection{Data collection}
A camera was set up approximately 5 meters from the robot's interaction location to video record individuals passing by or interacting with the robot. Additionally, researchers recorded observational notes on-site using audio to complement the video recordings. Conversations that passersby had with the robot were also audio recorded via the Zoom call that was set up to control the screen inside the robot. During the deployment, three researchers were on-site: two operated as wizards controlling the robots, while the third researcher provided support in observation, note-taking, and conducting interviews.

During the 15.5 hours of field study, we observed 29 instances where passersby~(p1–p29) intervened to help the robot resolve its uncertainties. Twelve of these interactions\footnote{An interaction was defined as any distinct occurrence where one or more passersby engaged with the robot to assist it in resolving a situation} involved individual participants, while the remaining 17 were group engagements. The rate of passerby involvement was relatively low given the average foot traffic and study duration at the three locations: 7.3 people/min over 6.5 hours at Location 1, 28.2 people/min over 5.5 hours at Location 2, and 5 people/min over 3.5 hours at Location 3 (see detailed distribution in Table~\ref{Overview}).

We approached and interviewed passersby who engaged with the robot, such as those who looked through the peephole or assisted the robot. Additionally, we interviewed passersby who drew their attention to the robot, for example, by pausing or gazing at it but eventually left without further interaction. The interviews included questions about participants' interactions with the robot and their opinions on intervening in the robot's operation as bystanders when the robot encounters uncertainties. Due to the nature of street interviews, some individuals hurriedly cut off or declined the interview request (n=5). As a result, we had to keep the interviews short and concise, ranging from 3 to 7 minutes (similar to comparable studies investigating interactions with robots in public spaces~\cite{Marius2020Woody, Hoggenmueller2020, Lee2019Bubble}). In total, we conducted 24 interviews with bystanders who engaged in interaction with the robot~(p1-p11, p13-p22, p24-p25, p27) and 21 interviews with those who left without further interaction~(c1-c21).

The study followed a protocol approved by our university’s human research ethics committee, and also complied with local regulations concerning data collection and the incidental recording of individuals in public spaces. We approached participants after their interactions with the robot to obtain their verbal consent for participation in interviews. Due to the rapid nature of street interviews, participants were not provided with an introduction to the Wizard of Oz method but received a printed participant information statement with further details on the study. In line with similar practices for field research in public spaces~\cite{Brown2024Public}, we did not seek consent from passersby who were incidentally included in the video recordings; however, measures were taken to protect their privacy by blurring identifiable features.

\subsection{Data analysis}
The data analysis employed a cross-analysis approach, where interview data provided deeper insights and explanations for passersby’s behaviours observed during the field study, while video recordings enriched the contextual understanding and supported the comments made during the interviews.
The first author closely examined the video recordings, documenting the incidents where passersby either helped the robot or noticed it without further interaction. These observations were captured through textual descriptions and supplemented by screenshots from the video. Subsequently, an approximately one-hour meeting was held between the three authors to discuss and review the interaction patterns identified by the first author during the analysis process.

To analyse the interview data, we transcribed the interview recordings and conducted a thematic analysis~\cite{braun2006thematic, Braun2019reflexive}. The first author examined and coded the data following an inductive approach, deriving sub-themes by identifying recurring patterns. This initial coding scheme was then discussed and refined during another one-hour meeting among three authors. Subsequently, the sub-themes derived from these codes were deductively grouped into final themes, structured around the two main aspects of our investigation: (1) supplementing interaction patterns identified from video analysis to interpret the reasons that influenced bystanders' behaviours and (2) understanding the impacts of engagement in the robot's uncertainty on bystanders' perceptions and attitudes toward the robot. This mixed approach aligns with \citet{Braun2019reflexive}'s reflexive thematic analysis, where they point out that coding could involve a blend of inductive and deductive processes rather than adhering strictly to one or the other.

\section{Results}
In this section, we first report common patterns of passersby's reactions, illustrated through representative instances derived from video analysis. We then supplemented our observations with interview data to interpret the triggers and barriers that influenced bystanders' involvement. Finally, we examine how bystanders' engagement during the robot's uncertainty impacts their perceptions and attitudes towards the robot, focusing on three key aspects: (1) maintaining human autonomy, (2) building connection through vulnerability and reciprocity, and (3) enhancing trust. 

\subsection{Passive observation without engagement}

\begin{figure*}[h]
\begin{center}
\includegraphics[width=1\textwidth]{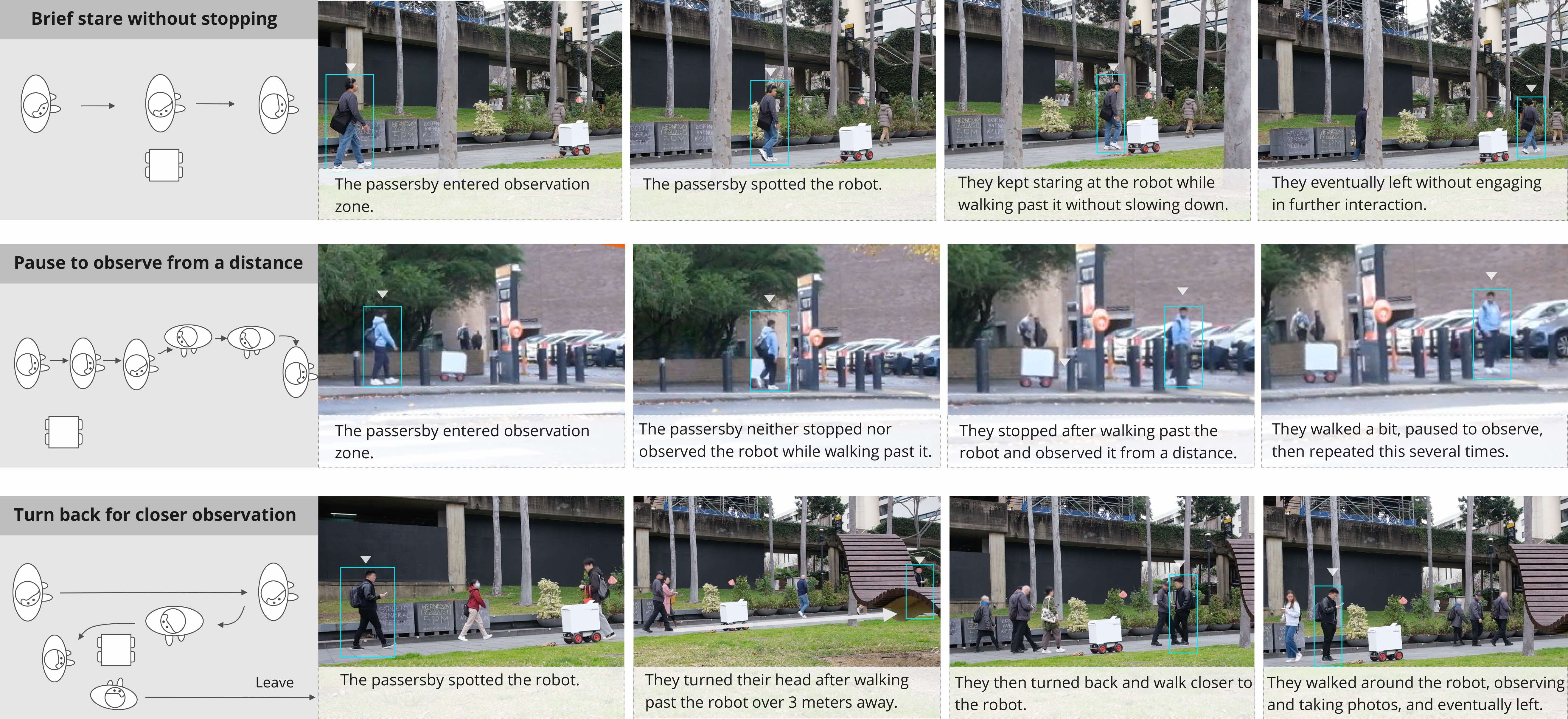}
\end{center}
\vspace{-8pt} %
\caption{Varying levels of attention displayed by passersby while remaining indifferent.}\label{Observe}
\Description{The image depicts three sections illustrating different levels of attention displayed by passersby towards the robot. Each section is displayed in a separate row, with each row containing a top-down view illustration on the left, showing the pedestrians’ trajectories around the robot, followed by a sequence of screenshots that depict the passerby's actions step by step.
Top row trajectory illustration labelled brief stare without stopping: The top-down view shows a pedestrian walking through the observation zone in a straight line, glancing briefly at the robot but not stopping. Screenshots sequences: (1) The image shows a passerby entering the observation zone from the left, with the robot ahead. The accompanying text reads: "The passersby entered the observation zone." ; (2) The passerby is walking closer to the robot and appears to have noticed it. The accompanying text reads: "The passersby spotted the robot." ; (3)The passerby continues walking past the robot while glancing at it, without slowing down. The accompanying text reads: "They kept staring at the robot while walking past it without slowing down."; (4) The passerby is leaving the observation zone, showing no further interaction with the robot. The accompanying text reads: "They eventually left without engaging in further interaction."

Middle row trajectory illustration labeled pause to observe from a distance: The top-down view shows a pedestrian walking through the observation zone, pausing briefly at a distance from the robot to observe it, and then resuming their path without direct engagement. Screenshot sequences: (1) The image shows a passerby entering the observation zone from the left, with the robot visible ahead. The accompanying text reads: "The passersby entered the observation zone."; (2) The passerby is walking past the robot without stopping. The accompanying text reads: "The passersby neither stopped nor observed the robot while walking past it."; (3) The passerby pauses at a distance, turning toward the robot to observe it. The accompanying text reads: "They stopped after walking past the robot and observed it from a distance."; (4) The passerby resumes walking, pauses again, and repeats this behavior multiple times. The accompanying text reads: "They walked a bit, paused to observe, then repeated this several times."

Bottom row trajectory illustration labeled turn back for closer observation: The top-down view shows a pedestrian walking past the robot in a straight line, then turning back after initially passing it, and approaching it more closely for detailed observation.
Screenshot sequences: (1) The image shows a passerby walking past the robot without stopping. The accompanying text reads: "The passersby spotted the robot."; (2) The passerby turns their head to look at the robot after walking past it by about 3 meters. The accompanying text reads: "They turned their head after walking past the robot over 3 meters away."; (3) The passerby turns back and starts walking toward the robot for a closer look. The accompanying text reads: "They then turned back and walked closer to the robot."; (4) The passerby walks around the robot, observing it from various angles and taking photos. The accompanying text reads: "They walked around the robot, observing and taking photos, and eventually left."}
\end{figure*}

\subsubsection{Low level of attention}
For much of the deployment, the majority of passersby walked past the robot without noticing it at all. Due to the commuter-focused nature of study locations 1 and 3, most people were singularly focused on reaching their destination, moving with purpose and not taking the time to look around. The robot did not receive a significantly higher level of attention after the animated flag pole was added. However, it managed to draw more attention after an audio cue was introduced, as we observed an increase in the number of passersby glancing at the robot due to the sound. This was further evidenced by instances where individuals who were originally focused on their phones shifted their attention to the robot after hearing the audio cue. Despite this increase in attention, the sound cue did not result in a significant increase in further engagement with the robot. 
\subsubsection{Remaining indifferent}
Among those who did notice the robot, a common interaction pattern was varying levels of attention—from brief stares to pausing for a longer observation—before eventually leaving without further engagement. Most passersby who noticed the robot would stare at it while continuing on their way, turning their heads to keep it in sight without slowing down, indicating an interest in observing the robot without a commitment to further engagement (see Fig.\ref{Observe}, top). Some passersby paused to observe the robot from a distance. For example, as shown in Fig.\ref{Observe}, middle, one passerby did not stop initially but paused after passing the robot to observe it. This individual repeated this pattern of walking a bit further, then pausing to observe again, several times before eventually leaving without further interaction. A smaller portion of passersby exhibited stronger interests in the robot by approaching the robot more closely, taking photos, or discussing it with companions, though they too eventually left without further engagement. As shown in Fig.\ref{Observe}, bottom, one passerby initially walked past the robot, then paused, turned back, and circled it for closer inspection. They then took a photograph of the robot before leaving. 
\subsubsection{Looking around and searching}
\begin{figure*}[h]
\begin{center}
\includegraphics[width=1\textwidth]{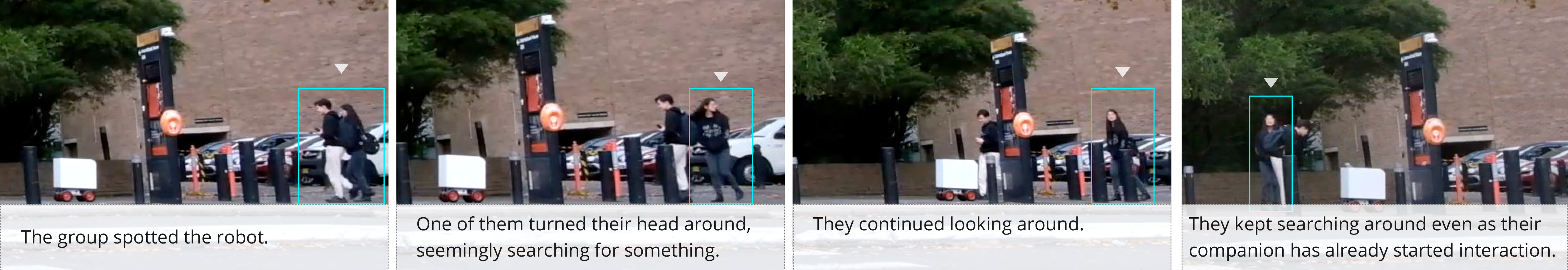}
\end{center}
\vspace{-8pt} %
\caption{Passersby looking around and searching after spotting the robot.}\label{Search}
\Description{This figure depicts a sequence of four screenshots capturing the actions of a group of passersby interacting with the robot. The images show the progression of their behavior after initially noticing the robot. Screenshot 1: The group notices the robot while walking through the observation zone. One passerby, highlighted in a blue bounding box, appears to have spotted the robot first. The accompanying text reads: "The group spotted the robot." Screenshot 2: One of the group members turns their head, seemingly scanning the surroundings for additional context or information. The accompanying text reads: "One of them turned their head around, seemingly searching for something." Screenshot 3: The group continues to look around, maintaining some focus on the robot but not engaging directly yet. The accompanying text reads: "They continued looking around." Screenshot 4: While one group member begins interacting with the robot, another continues to search the area, indicating differing levels of engagement. The accompanying text reads: "They kept searching around even as their companion has already started interaction."}
\end{figure*}

Another noticeable observation was that some passersby would look around, seemingly searching for something after spotting the robot, which often led to their eventual departure without further interaction. In Fig.\ref{Search}, one passerby can be seen turning their head repeatedly, scanning the surroundings, even as their companion began interacting with the robot. In the interview, this participant explained that they believed the robot was being tested and were looking for the person in charge of it. This observation was particularly pronounced during the first few days of the field study at Location 1, where foot traffic was relatively low, even leading to instances where the Wizards were spotted during the pilot run.

\subsection{Getting involved and offer help}
\begin{figure*}[h]
\begin{center}
\includegraphics[width=1\textwidth]{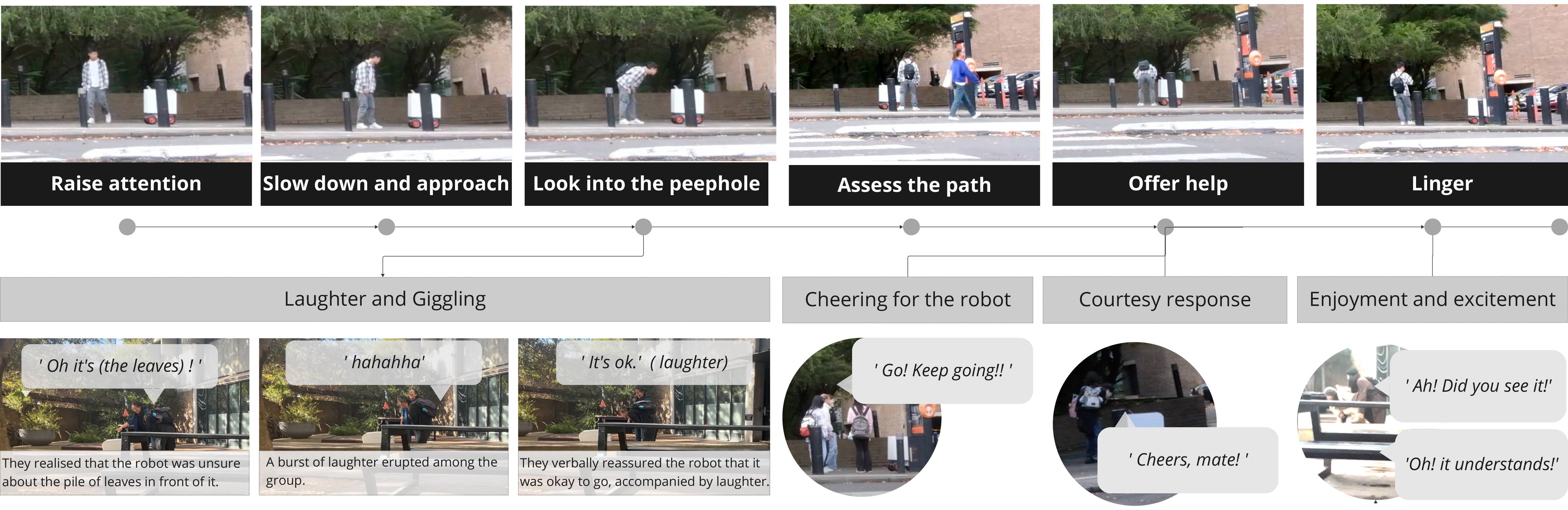}
\end{center}
\vspace{-8pt} %
\caption{Interaction sequence of a passerby getting involved and resolving the robot's uncertainty, with additional observations supported by supplementary images and quotations.}\label{Involve}
\Description{The image illustrates a sequence of passersby interacting with a robot, including six stages: (1) Raise attention: The passerby notices the robot. (2) Slow down and approach: They slow down and move closer. (3) Look into the peephole: The passerby bends over to peek inside the robot. (4) Assess the path: They observe the surroundings, considering the robot's uncertainty. (5) Offer help: The passerby offers assistance.(6) Linger: They stay after helping. Below the sequence, there are various conversational reactions, such as laughter, cheering, and excitement.
}
\end{figure*}

Among the 29 interaction instances where passersby intervened to help the robot resolve its situation, 27 involved looking into the peephole, reading the information about the robot's uncertainty, and offering help accordingly. In two instances, the participants did not look into the peephole; instead, they inferred the situation from the context and helped the robot by clearing the path. Fig.\ref{Involve} illustrates a typical interaction sequence of how a passerby becomes involved in helping the robot navigate an uncertain situation. %Additional observations from other instances were incorporated into the overall process, supported by supplementary images and quotations.

\subsubsection{Reactions to uncertainty: laughter and giggling}
After noticing the robot, passersby slowed down, approached it, and bent down to look through the peephole. Upon seeing the screen and reading about the robot's uncertainty information, a common reaction among people in groups was laughter and giggling, sometimes sparking discussions within groups, describing the robot as \emph{`So Cute!'}. As shown in Fig.\ref{Involve}, bottom, the group erupted in laughter when they realised the robot was unsure about the pile of leaves and puddle in front of it. They later responded to the robot, saying \emph{`it's okay,'} while their laughter continued. This sense of enjoyment and delight, particularly when discovering the robot’s limitations, was articulated by p18 as being \emph{`pleasantly surprised'}. They further described the experience as akin to \emph{`finding a gem outside, you didn't expect it'}. 

\subsubsection{Offering help}
Upon understanding the robot's uncertainty, participants commonly exhibited a behaviour of briefly examining the obstacle and assessing the path ahead before offering any response. To respond to the robot, the majority (n=23) verbally communicated with it, often saying things like \emph{`Yeah, just keep going'} or \emph{`Yes, it's safe to go'} to indicate it was safe to proceed. The utterances were generally short and precise. We also observed groups cheering for the robot, saying things like \emph{`Go! Keep Going!'} in an encouraging tone. Additionally, three participants demonstrated social etiquette by responding to the robot’s \emph{`Thank you!'}---displayed on the screen by the wizard---with phrases like \emph{`Cheers, mate!'}~(p1) or \emph{`You're welcome.'}~(p21)

Other than verbal response, four people attempted to resolve the robot's uncertainty by physically clearing the path ahead. For example, after checking the screen, p19 walked around to the front of the robot, cleared the path, and then returned to the peephole, saying to the robot \emph{`Yeah, there are no more leaves'}. Two participants directly pushed the robot forward after assessing that the path ahead was safe. 

\subsubsection{Enjoyment and excitement after interaction}
After giving the robot instructions, people tended to linger, eagerly anticipating how the robot would respond to their input. When the robot resumed its movement~(initiated by the wizard), it often sparked genuine surprise and excitement, especially in groups where people's reactions were more animated. People would exclaim \emph{`Wow!'}, laugh, or turn to their friends with joyful astonishment, saying things like \emph{`Did you see that?!'} or \emph{`Oh, it understands!'} as they marvelled at the robot's responsiveness (see Fig.\ref{Involve}, bottom-right). 

The observed enjoyment and excitement were supported by subsequent interviews, where the majority of participants reported positive experiences when engaging with the robot during its moments of uncertainty~(n=13). Some participants described the experience as \emph{`fun'}~(p15) or \emph{`interesting'}~(p8). However, one participant expressed a negative view, stating \emph{`I don't like that'}, as they believed that \emph{`[…] robots, you know, by definition, should be very binary—right on or off,'} rather than displaying uncertainties and requiring human assistance.

\subsection{Engagement barriers and triggers}

\subsubsection{Barriers to engagement} The level of indifference and low engagement was unexpected, as at the time of the study, no mobile robots were deployed on the university campus or in the broader region where the campus is located. Given this novelty, we anticipated that the robot would spark greater interest and a higher level of involvement from passersby. 

To understand the reasons for this, we interviewed 21 passersby 
%\footnote{We refer to these passersby we interviewed who did not interact with the robot as c1–c21, to distinguish them from participants we interviewed who had interactions with the robots.}
who noticed the robot (identified by observing their gaze towards it) but eventually moved on. Nine passersby indicated that they were unaware of the robot's need for assistance and continued on their way due to the robot's implicit signalling of its need for help. Two of them misinterpreted the robot's flagpole waving as a mere \emph{`greeting'} rather than a request for help. Additionally, three passersby described the beep sound as simply drawing attention, viewing it as a \emph{`warning to avoid bumping into it'} rather than an indication that the robot required assistance. Three participants indicated that they were in a rush and didn't pay much attention to the robot.

The remaining 9 passersby, despite understanding the robot was in a difficult situation, remained indifferent. They inferred from the flag motion, beep sound, and the robot's lack of movement that it might be stuck, yet chose not to engage. Five of these individuals mentioned that their indifference stemmed from the belief that the robot was someone else's property, and therefore, they felt they were not meant to interact with it. For example, c13 suggested that they \emph{`thought it would just be rude to touch it'}. Two of them further expressed concerns about \emph{`break[ing]'} the robot, which \emph{`looks expensive'}, if they got involved in interacting with it. In addition, three passersby indicated that they felt unqualified to intervene in such situations, believing that professional assistance might be required. As exemplified by c10, who perceived the robot's \emph{`flag waving around'} as not being directed at them, but \emph{`some uni staff would be called there'} instead. 

\subsubsection{Triggers for engagement}
As a bystander, completing the interaction sequence of assisting the robot involves two key motivational steps. First, they must be motivated to approach the robot and look into the peephole to assess the information about the robot's uncertainty. Second, they need to be motivated to offer help based on the information. 

\emph{Curiosity and impacts of additional cues. }Before accessing the information about the robot's uncertainty hidden behind the peephole, curiosity emerged as the primary motivating factor that led passersby to pause and approach the robot for closer observation~(n=14). Six participants linked this curiosity directly to the peephole, as p18 noting that \emph{`a combination of curiosity and fun-seeking nature'} led them to engage further by looking inside. Five participants suggested that the robot itself served as a trigger. Their motivation for getting closer to observe the robot was simply because they \emph{`saw a large white box'}~(p3) or believed the robot \emph{`was stuck'}~(p13). After the animated flag and audio cue were added, 12 participants mentioned that these cues attracted their attention and prompted them to stop and observe the robot. However, most of them (n=9) indicated they did not derive any underlying meaning of help-seeking from the cues beyond their attention-grabbing function. It is worth noting that participants only began to infer that the gestural cue was a request for help after the audio cues were added. As participants p24 and p25 specifically pointed out, it was the combination of the beep and the gesture that made them understand the robot's intent.

\emph{Care and contribution to robot learning. }Once bystanders read the screen and understood that the robot was stuck due to uncertainty, some participants indicated that their helping behaviours were driven by empathy; for example, one participant stated they \emph{`felt bad for it'}~(p8). P1 further explained that this sense of care could be attributed to the peeping interaction, which brought them physically closer to the robot and lowered their height, fostering a more intimate engagement. P1 likened this mode of interaction to engaging with \emph{`a puppy'}, which evoked a nurturing and caring response. Moreover, the sense of care can be attributed not only to the robot itself but also extended to the person associated with it, such as the service recipient or the robot's owner~(n=2). For example, p17 suggested that their reason for offering help was out of concern that \emph{`someone's delivery is going to be late'}. Beyond care, four participants mentioned that their motivation stemmed from the belief that their assistance could contribute to the learning and training of the algorithm driving the robot, thereby helping to improve its operational capabilities. As p2 suggested, \emph{`Because it helps the robot to learn, so it can evolve faster.'} 

\begin{figure*}[h]
\begin{center}
\includegraphics[width=1\textwidth]{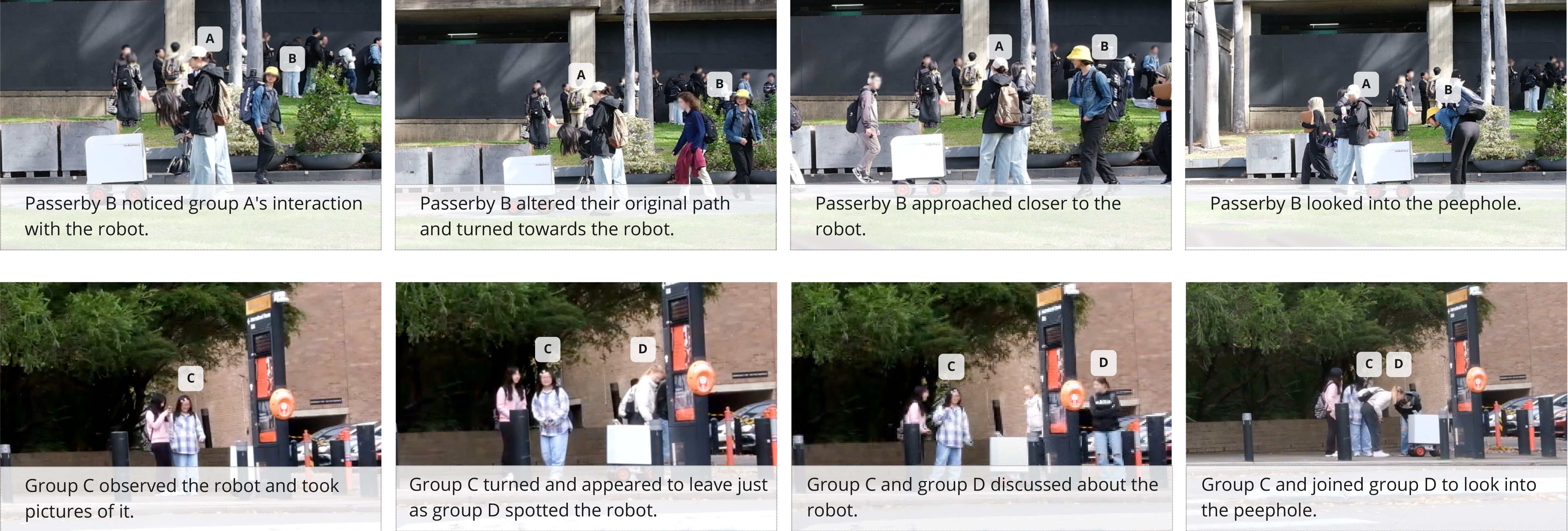}
\end{center}
\vspace{-8pt} %
\caption{Social influence motivating bystander engagement: (Top) Passerby B approaches and interacts with the robot after observing Group A; (Bottom) Group C and Group D collectively engage with the robot after initial hesitation.}\label{Influenced}
\Description{This figure illustrates how social dynamics influence passersby’s interactions with the robot, presenting sequences of actions in two rows, each with four screenshots. The top row depicts Passerby B’s engagement with the robot after noticing Group A’s interaction. In Screenshot 1, Passerby B notices Group A standing near the robot, drawing their attention. The accompanying text reads: "Passerby B noticed Group A's interaction with the robot." In Screenshot 2, Passerby B alters their original path and turns toward the robot, showing curiosity and interest. The accompanying text reads: "Passerby B altered their original path and turned towards the robot." In Screenshot 3, Passerby B moves closer to the robot, increasing their engagement. The accompanying text reads: "Passerby B approached closer to the robot." In Screenshot 4, Passerby B leans forward to look into the robot’s peephole, engaging directly. The accompanying text reads: "Passerby B looked into the peephole."

The bottom row illustrates Group C and Group D’s engagement with the robot, demonstrating the influence of social dynamics. In Screenshot 5, Group C observes the robot from a distance and takes pictures of it, showing initial interest. The accompanying text reads: "Group C observed the robot and took pictures of it." In Screenshot 6, Group C starts to leave the area as Group D notices the robot, reflecting overlapping social dynamics. The accompanying text reads: "Group C turned and appeared to leave just as Group D spotted the robot." In Screenshot 7, Group C and Group D engage in a discussion about the robot, sharing observations and becoming collectively interested. The accompanying text reads: "Group C and Group D discussed about the robot." In Screenshot 8, Group C and Group D join together to look into the robot’s peephole, demonstrating a joint interaction and deeper engagement. The accompanying text reads: "Group C and Group D joined to look into the peephole."}
\end{figure*}

\emph{Social influence. }In addition to instances where bystanders engaged with the robot on their own initiative, we observed cases where passersby were influenced by the actions of others, which subsequently prompted further engagement. The interaction between bystanders and the robot often served as a catalyst, drawing additional attention from nearby individuals. For example, as shown in Fig. \ref{Influenced}~(top), passerby B noticed group A interacting with the robot. This observation led passerby B to alter their path, approach the robot, and engage with it by looking into the peephole after group A left. Additionally, we observed that social influence played a role in encouraging further engagement with the robot, particularly in cases where hesitant passersby were influenced by others to interact. As depicted in Fig.\ref{Influenced}~(bottom), Group C initially stopped by the robot, engaging in observation and discussion. However, they hesitated to interact further. As they were about to leave, another group of passersby noticed the robot, prompting a conversation between the two groups. This exchange ultimately led both groups to collectively engage with the robot. 

\subsection{Impacts on bystander's attitudes towards robot}
In this section, we report insights from post-interaction interviews on how involvement in the robot's uncertainty impacts people's perceptions and attitudes towards robots.

\subsubsection{Human autonomy and control} 

We observed a trend in the interviews where participants seemed to feel empowered by maintaining control over the robot's autonomy, which may have contributed to these positive views. Nine participants expressed satisfaction they derived from assisting the robot having uncertainties, describing how it made them feel \emph{`[the robot] being depended on [them]'}~(p5, p9), \emph{`important'}~(p15), \emph{`powerful'}~(p22), and even \emph{`superior to the robot'}~(p22). Four participants, when discussing their intervention, interestingly remarked that as an indication that robots are yet not dominating, as p19 stated \emph{`[it] shows me that they are not here taking over the world'} and \emph{`it's a good sign that I'm still valuable, you know, as a human'}. Two participants expressed a preference for robots involving bystanders when facing uncertainties over fully autonomous robots, as they enjoyed the feeling of having \emph{`control over it'}~(p5), which even made them feel \emph{`safer'} around the robot.

\subsubsection{Connection through vulnerability and reciprocity} 

Five participants tended to anthropomorphise the robot when it communicated uncertainties, describing it as vulnerable entities, such as a \emph{`kid'}~(p15,16,18) or a \emph{`little dog'}~(p21) in need of help. Such perceived vulnerability contrasted with the typical stereotype of robots as perfect and infallible, instead highlighting more human traits. Four participants indicated that the robot seemed \emph{`more human'} by acknowledging that these machines are not \emph{`always correct'} but can \emph{`also make some faults'}~(p17). Two participants further noted that this vulnerability made the robot more \emph{`approachable'} and \emph{`affable'}. 

Engaging with the robot during moments of uncertainty helped reduce the perceived distance between bystanders and the robot, fostering a sense of connection. As p6 stated, \emph{`[…] normally for robots that operate autonomously, as a bystander, you normally don't have anywhere to interact with it'}, and they \emph{`felt good […] with this one, because it actually approached you and wanted some help.'} Additionally, five participants reported feeling \emph{`closer to it (the robot)'}~(p1) due to their interactions with the robot. P18 further described developing a \emph{`personal connection'} with the robot, even as a bystander, attributing this to the casual collaboration that spontaneously formed during their encounter. Three participants mentioned the reciprocal nature of these casual collaborations, acknowledging that while the robot provides services to humans, a more \emph{`symbiotic'}~(p5) relationship is formed if humans also offer assistance when needed.

\subsubsection{Trust}
Although the objects that we staged for the robot to have trouble recognising were generally perceived as easy to identify, only one participant reported a decrease in their trust towards the robot. In contrast, six participants indicated that the robot's uncertainties did not significantly impact their trust. Interestingly, ten participants reported attributing even greater trust to the robot. This increase in trust was likely attributed to the perceived intelligence of the robot, as it was capable of acknowledging uncertainty and appropriately seeking assistance. As p11 suggested, they felt that the robot \emph{`has the ability of independent thinking'}. Furthermore, five participants considered the robot’s search for information during uncertainty as an essential part of its training process, enhancing its technological capabilities. This belief could also contribute to increased trust, as people perceived the robot as actively improving its performance. As p3 noted, \emph{`it's trying to learn, like it's trying to get more data […] once we give it an input, next time, when it faces the same problem, it can probably resolve it itself'}. 

In addition to the perceived intelligence and belief in enhancing the robot's technological capabilities, the notion of \emph{`being part of the process'}~(p5) and thereby maintaining human autonomy in the robot's operation emerged as another reason for attributing trust, as indicated by three participants. The ability to intervene helped mitigate concerns about technology dominating, which could further enhance trust. As p19 suggested, \emph{`I can see that it's not completely taking over the world. So it shows that the robot still needs the human side, and that improves the trust relationship'}.

\section{Discussion}
Drawing on the findings of the previous section, we first discuss the potential benefits and issues of involving bystanders in mitigating urban robots' technological imperfections. Reflecting on our design concepts, we further derive design implications for implicit and non-intrusive engagement strategies that encourage casual human-robot collaboration, supporting bystander participation in mitigating technological imperfections. These implications include \emph{balancing bystander autonomy and persuasiveness}, \emph{incorporating intentional friction to enhance engagement}, and \emph{utilising robot gestural cues in public spaces}.

\subsection{Involving bystanders in mitigating urban robots' technological imperfections}
In the current debates regarding the impact of AI technology on human society, the \emph{technology-centric view}~\cite{peeters2021hybrid} suggests that AI will soon outperform humankind and eventually take over, potentially becoming the primary threat to humanity. Such concerns have influenced people's attitudes towards robots, leading to tensions about the introduction of robots into public spaces~\cite{Han2023InourStreet, Bennett2021CrowdedSidewalk}. A notable example occurred in San Francisco, where residents called for a ban on urban delivery robots shortly after their deployment~\cite{TheGuardian2017DeliveryRobots}, driven by not only concerns about public safety, but also job displacement due to implications of automation.

This apprehension was also reflected in our field study, where some participants expressed concerns about fully independent robots in the interview. Interestingly, they indicated a sense of relief upon noticing the robot's uncertainty, viewing it as an indication that robots are not yet capable of total autonomy. They noted that the robot showing uncertainty demonstrated vulnerability, reassuring them that robots are not yet \emph{`taking over the future'}~(p5). By assisting the robot in navigating its uncertainty, participants felt empowered, reinforcing their sense of human superiority, as p22 described feeling \emph{`powerful'} and \emph{`superior to the robot'}.

While a robot's competence is often identified as a factor influencing people's social perceptions~\cite{Carpinella2017RoSAS} and trust~\cite{christoforakos2021competenceTrust} towards robots, our study revealed a different pattern. Instead of diminishing trust, nearly half of the participants reported an increase in trust towards the robot, despite it being stuck due to uncertainty. This can be partially attributed to the fact that bystanders were not direct service recipients, so the robot’s uncertainty had little impact on them. More importantly, the increase in trust was likely tied to the autonomy and control participants maintained during the interaction. Keeping humans in the loop, even through simple actions like pressing a button to approve the robot’s plan execution, has been shown to significantly enhance trust in the robot~\cite{Ullman2017Button}.

Additionally, passersby's reactions of laughter and delight upon discovering the robot's uncertainty suggest their affection towards what people perceived as the robot expressing vulnerability, which they described as \emph{`endearing'}~(p19) and \emph{`cute'}~(p20). This affection is similar to previous findings in social robots, where people significantly preferred a robot making mistakes over a flawless one~\cite{Nicole2017Err}. While this study focused on humanoid robots with social capabilities and involved humans as collaborators, our findings extend this understanding. We show that affection towards a robot's mistakes can also manifest with a service robot designed for task execution rather than sociability, even in public spaces where bystanders have no prior relationship with the robot.{}

Our study demonstrates that the positive impact of maintaining human autonomy and allowing robots to display technological imperfections can extend beyond traditional human-robot collaboration settings~\cite{SarahCHI2024Uncertainty,Hough2017RobotUncertainy,Moon2021Hesitation} to public contexts, where humans engage as bystanders rather than collaborators. While commercially deployed urban robots increasingly come equipped with help-seeking features to request human assistance when facing physical limitations (e.g., getting immobilised due to obstructed road conditions)~\cite{Boos2022Polite}, these features often do not extend to situations involving technological imperfections, such as uncertainties in navigation decision-making. In such cases, robots still rely on remote human supervisors hidden in control rooms to resolve their uncertainties~\cite{shaw2022california}, concealing the robots' struggles from the public. Our findings highlight the opportunity to engage bystanders' support in these moments of technological imperfection. This approach not only addresses the robots' operational challenges but also has the potential to alleviate public concerns (e.g., fears of technological domination) and foster closer relationships between urban robots and bystanders, ultimately promoting smoother integration of urban robots into everyday life. At the same time, however, exposing the robot’s vulnerabilities could inadvertently place an unfair burden on the public, raising ethical concerns about the expectation of free labour~\cite{hakli2023helpingAsWorkCare}. Furthermore, relying on bystander input introduces the risk of unreliable assistance, necessitating appropriate oversight mechanisms to ensure that bystander interventions do not compromise the robot’s functionality.

\subsection{Balancing bystander autonomy and persuasiveness}
One noteworthy observation that emerged from our study was the low dropout rate once bystanders initiated interaction by peeking into the peephole, despite the overall limited engagement relative to foot traffic. In our study, only two out of 29 passersby who peeked into the peephole disengaged without offering further assistance. This stands in contrast to the study by \citet{Weiss2015Direct}, which investigated a similar scenario where a robot explicitly requested directions from passersby through verbal communication. In their study, out of approximately 100 robot-initiated interactions, only 36 participants completed the interaction by providing directions. Unlike their study, where participants were prompted by the robot's explicit verbal help-seeking request, the initiative taken by passersby in our study was rooted in intrinsic motivation, driven by human curiosity and playfulness in `peeking'. This contrast illustrates the principles of ludic engagement, where the peephole's openness and ambiguity subtly invite bystanders' exploration without explicitly compelling engagement, leading to deeper and more self-motivated involvement.

Furthermore, this can be understood through the lens of self-determination theory~\cite{Ballou2022SDT, ryan2018SDT}, which suggests that when individuals initiate actions themselves, they experience enhanced autonomy, leading to deeper intrinsic motivation. As a result, they are more likely to commit to continuing the interaction due to a sense of ownership over the action. This theory has been applied in HCI to address interaction motivation across domains such as learning~\cite{Dhiman2024SDT}, game design~\cite{Tyack2020SDTGame}, and human-robot interaction~\cite{van2020SDT}.

While bystanders were more likely to continue their involvement and offer help when they took the initiative, our study also reveals barriers for bystanders in initiating interactions with a robot they had no pre-existing relationship with, as reflected in the overall limited engagement relative to foot traffic. Even when they had the intention to interact, bystanders often hesitated, perceiving the robot as someone else's property rather than something that they could freely engage with. 
This perception may account for our observation of passersby often looking around, seemingly searching for a person `in charge' when they noticed the robot in need of assistance. This suggests that before bystanders can transition from passive observers to active collaborators, the robot must provide a clear signal of openness, essentially granting permission for further interaction.

The barriers led us towards design decisions of adding additional cues to `nudge' bystanders into engaging with and helping the robot. This effort to enhance engagement surfaced a tension: our aim to uphold bystander autonomy through non-intrusive design conflicted with the potential risk of crossing into manipulative territory. This tension resonates with broader concerns in HCI regarding the use of design power~\cite{Kender2022DesignPower} and the implementation of dark patterns~\cite{Lace2019DarkpatternRobotCuteness, Mathur2021darkpattern} to manipulate user behaviours. Such concerns become especially pertinent in the context of bystanders assisting commercially deployed robots, where the persuasiveness of these design strategies could lead to issues of invisible labour and potential exploitation~\cite{hakli2023helpingAsWorkCare}. Thus, careful consideration must be given to balancing bystander autonomy and the persuasiveness of engagement strategies.

\subsection{Incorporating intentional friction to enhance engagement}

Unlike common interaction design principles that tend to emphasise ease of use~\cite{rogers2011interaction}, our design concept introduced a level of friction, requiring passersby to bend over or even squat to peek through a small peephole to access information and proceed with the interaction. The low dropout rate suggests that these physical barriers did not deter passersby from further engagement; instead, the added physical challenge unexpectedly appeared to enhance their sense of involvement. This aligns with the concept of psychological ownership~\cite{Pierce2003PsychologicalOwnership}, which suggests that effort invested in an object or task increases individuals' sense of ownership and personal connection, motivating deeper engagement and commitment. The observation of bystanders lingering and waiting for the robot to resume movement further demonstrates this commitment, suggesting that their continued presence reflects a sense of care~\cite{hakli2023helpingAsWorkCare} developed through the interaction.

Design frictions refer to points of difficulty encountered during a user's interaction with technology, which are often minimised to increase and sustain user engagement with a product~\cite{Cox2016Friction}. However, \citet{Cox2016Friction} argue that incorporating friction can prompt reflection and foster more mindful interaction. This approach has been used to reduce input errors~\cite{Sarah2013number} and promote behaviour change, such as reducing smartphone distractions~\cite{Dutt2024Smartphone}. While design friction is often used to discourage certain actions, our study shows it can also be utilised to motivate deeper bystander involvement by increasing their physical investment in interactions. When designing for voluntary, casual human-robot collaboration, designers could consider introducing intentional friction into interactions, where the challenge itself becomes a driver for continued involvement.

\subsection{Utilising robot gestural cues in public spaces}
The introduction of gestural cues (i.e., the animated flagpole) did not significantly increase the number of passersby who paid attention, nor did it raise the percentage of individuals who paused for further engagement. While previous research suggests that humans tend to perceive non-verbal gestures from non-humanoid robotic objects as social signals~\cite{Novikova2014DesignModel,Erel2022emotionSupport, Erel2024OpeningEncounter,Press2022HumorousGestures, Erel2024ObjectinAV} that invite further interactions~\cite{ju2009approachability,sirkin2015Ottoman}, our study revealed gaps in transferring this effectiveness to public spaces. This could be explained by the contextual complexity of public spaces and the casual nature of the setting, where people without a pre-determined intention to interact may not interpret such cues as invitations for further interaction.
% in a public space setting, as well as the casual situation with people not having a pre-determined intention to interact with a robot hinders the interpretation of such cues as a interaction for further interaction .

Passersby’s attention is often fragmented. Based on our observations, most passersby offered only brief attention to the robot, often limited to a quick glance, making it difficult for them to grasp the full context of a motion sequence and infer its underlying meaning. In addition, previous studies of gestural cues were often conducted in controlled lab settings where participants were primed with context to observe and interpret the robot’s behaviour. In contrast, random passersby in public spaces lack this contextual framing, making it difficult for them to interpret the meaning of the robot's gestures. The study most comparable to ours is~\cite{ju2009approachability}, where authors observed passersby’s spontaneous reactions to an automated door's gestures (e.g., different movement trajectories) in a field setting. The study found significant uniformity in participants' interpretations, perceiving the door's gestures as conveying different approachability. However, in this study, passersby intended to pass through the automated door, making attention to the door's gestures an inevitable step before they could proceed with their primary goal. In contrast, in our study, the robot's gestural cues were completely irrelevant to the activities that bystanders were engaged in. 

Interestingly, after the auditory cue was added, a noticeable shift occurred: participants began to infer that the gestural cues were communicating a need for help, interpreting this through the combination of auditory and gestural signals. This contrasts with participants' perception of the animated flagpole, which, prior to the audio cue being added, was seen as having no communicative intent beyond greeting or attracting attention.

Our findings suggest that passersby's fragmented attention and the lack of contextual framing in public spaces can make the nuances in robot gestural cues difficult to grasp. Designers should consider limiting the complexity of such cues to basic functions, such as attracting attention or supplementing them with other signals, such as audio, to provide more contextual information and enhance communication effectiveness.

\subsection{Limitations}
First, our field study took place on a university campus, where students comprised the majority of passersby. This context may have influenced the nature of interactions, limiting the transferability of our findings to other public spaces with different demographic compositions. Second, the evolving nature of our design concept and the variation in deployment locations introduced complexities that shaped our observations and insights, making it challenging to achieve traditional data saturation ~\cite{Virginia2021saturate}. This is due to our study following an RtD approach ~\cite{Zimmerman2007RtD}, which emphasises iterative, exploratory processes where insights evolve throughout the research. This epistemological stance shifts the focus from traditional saturation to capturing the reflectiveness and relevance of insights, which we ensured by meticulously documenting the evolution of our design concepts and articulating the insights generated throughout the iterative process and field study.

\section{Conclusion}
This study approached the challenge of urban robot uncertainty from a novel angle, moving beyond purely technological advancements to actively engage bystanders as participants in mitigating robot uncertainty. We designed a speculative \emph{peephole} concept that invites bystanders to help resolve the robot's uncertainty in a non-intrusive, curiosity-driven manner. The concept was tested in a field study to probe passersby's spontaneous reactions and examine how their involvement shaped their perceptions and attitudes towards the robot. Despite the low overall engagement relative to the foot traffic, we observed a strikingly low dropout rate among those who took the initiative to engage with the peephole, with many expressing genuine enjoyment from the interaction. At the same time, a sense of empowerment, affection, and connection towards the robot emerged as bystanders helped the robot resolve its uncertainty.

Drawing on these findings, we highlight the potential of involving bystanders to mitigate urban robots’ technological imperfections, which can not only address operational challenges but also foster public acceptance of robots. Furthermore, our reflections on the design concepts offer practical design implications that designers can leverage to support bystander engagement with urban robot technological imperfections. These include balancing bystander autonomy and persuasiveness, incorporating intentional friction to encourage deeper engagement, and the use of robot gestural cues in public spaces. While speculative in nature, our concept served as a vehicle for knowledge generation, providing insights on how to engage bystanders in urban robot uncertainty while minimising intrusiveness. We hope that our work lays a valuable foundation for future studies to build on these implications and develop practical solutions, facilitating the seamless integration of urban robots into everyday life and fostering symbiotic human-robot relationships.

\begin{acks}
This study is funded by the Australian Research Council through the ARC Discovery Project DP220102019 Shared-Space Interactions Between People and Autonomous Vehicles. We thank all the participants for taking part in this research. We also thank the anonymous CHI’25 reviewers and ACs for their constructive feedback and suggestions to make this contribution stronger.
\end{acks}

\bibliographystyle{ACM-Reference-Format}
\bibliography{sample-base}


\end{document}